\documentclass[conference]{IEEEtran}
\usepackage{cite}
\usepackage{amsmath,amssymb,amsfonts}
\usepackage{algorithmic}
\usepackage{graphicx}
\usepackage{textcomp}
\usepackage{xcolor}
\def\BibTeX{{\rm B\kern-.05em{\sc i\kern-.025em b}\kern-.08em
    T\kern-.1667em\lower.7ex\hbox{E}\kern-.125emX}}
\usepackage{graphicx}
\usepackage{subcaption}
\usepackage{float}

\newif\ifcomment
\commenttrue

\usepackage[numbers]{natbib}
\begin{document}

\title{Dynamic Task Allocation with Dialogue}

\author{
\IEEEauthorblockN{Matthew Groholski, Weiwei Gu, and Nakul Gopalan}
\IEEEauthorblockA{
School of Computing and Augmented Intelligence \\
Arizona State University \\
Tempe, AZ, USA \\
\{mgrohols, weiweigu, ngopala6\}@asu.edu
}
}

\maketitle

\begin{abstract}
Effective multi-agent collaboration requires agents to adapt task allocation as goals and responsibilities evolve. 
Prior task-allocation frameworks often assume static goals, full observability, or rely on large language models~(LLMs) without explicit, verifiable beliefs about task allocation or intent. 
These assumptions limit their applicability to real-world collaboration, where agents often lack full information about each other's actions, progress, or local environment.
To support dynamic task allocation under partial observability, we introduce a communication-augmented Bayesian task-allocation framework.
Our agent uses dialogue to reduce uncertainty about the environment and collaborators, while keeping decisions grounded in explicit probabilistic beliefs. 
We evaluate our framework in an Overcooked domain extended with dynamically arriving tasks under both full and partial object-space observability.
Our communication agent consistently outperforms a vanilla LLM agent in both settings.
Additionally, our framework performs comparably to a non-communicating task-allocation baseline under full observability, but substantially outperforms it under partial observability as task complexity increases by an average speed up of 0.585 and completes on average 18 timesteps faster. 
These results indicate that our agent is more efficient in dynamic task allocation than baselines in partially observable environments, and that communication is important as available information decreases.

\end{abstract}


\section{Introduction}
\begin{figure*}[htbp]
\centering
\begin{subfigure}[b]{0.32\linewidth}
\centering
\includegraphics[width=\linewidth]{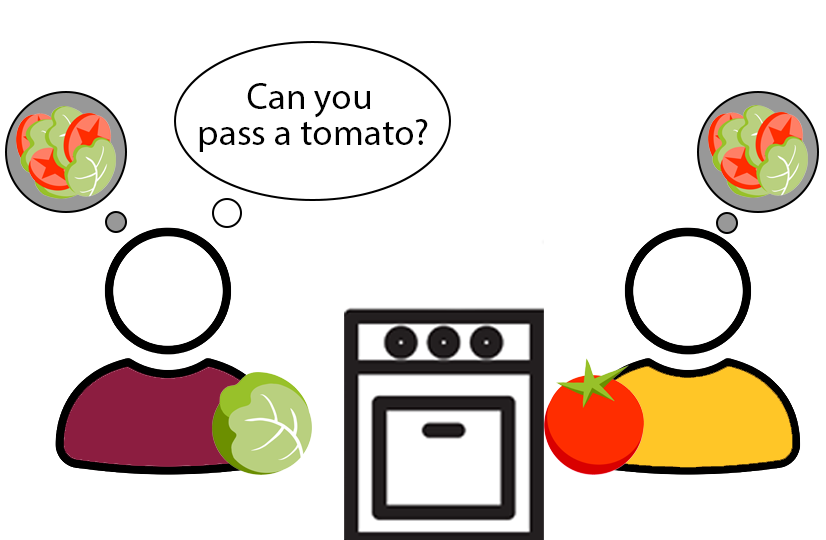}
\caption{Timestep 1}
\end{subfigure}
\hfill
\begin{subfigure}[b]{0.32\linewidth}
\centering
\includegraphics[width=\linewidth]{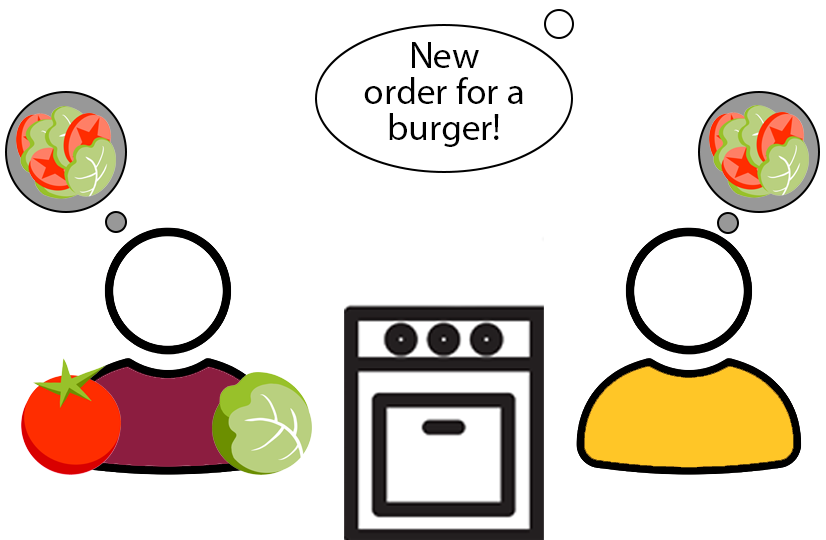}
\caption{Timestep 2}
\end{subfigure}
\begin{subfigure}[b]{0.32\linewidth}
\centering
\includegraphics[width=\linewidth]{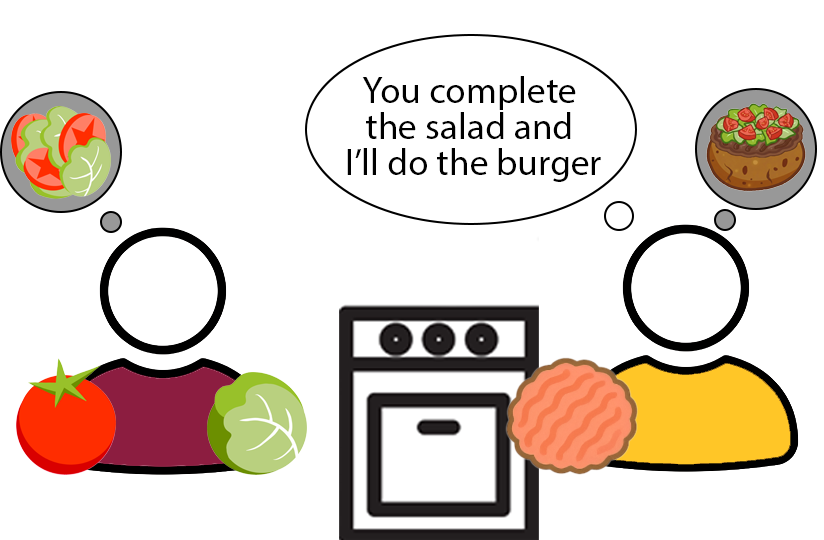}
\caption{Timestep 3}
\end{subfigure}

\caption{Two humans communicating to coordinate a shared goal, with each human's current belief about their assigned task shown in the gray thought bubble. (a) The human in maroon asks the human in gold to pass a tomato while both are working on the same salad task. (b) A new order for a burger arrives. (c) The human in gold reassigns responsibilities to divide the work more efficiently, delegating the salad to the human in maroon and taking on the burger itself.}
\label{fig:intro:example}
\end{figure*}

% \todo{Shorten paragraph}
Humans often collaborate in environments where goals, responsibilities, and progress constantly evolve, from cooking in a kitchen to performing symphonies~\citep{melis2010how-5ab}.
In these settings, collaborators must continually coordinate how duties are distributed and updated over time, a process known as \textit{dynamic task allocation}.
In a busy kitchen, for example, cooks redistribute tasks such as preparation, cooking, plating, and serving based on workload and individual progress.
Dialog is one of the most common modalities for this coordination, enabling collaborators to communicate progress, negotiate responsibilities, and maintain a shared understanding of the task~\cite{klimoski1994team-4d4, clark1996using, griffiths2015rational, lieder2020resource}, as shown in Figure~\ref{fig:intro:example}.
Effective human-agent collaboration therefore requires artificial agents that can use dialog to perform dynamic task allocation, reasoning about collaborators' goals, actions, and progress to adapt responsibilities as an interaction unfolds~\cite{nikolaidis2012humanrobot-d73}.

Prior work has used reinforcement learning and large language models~(LLMs) to develop collaborative agents~\citep{collins2024building-5f4, unhelkar2020decisionmaking-7d5, liu2023llmpowered-786, sun2025collabovercooked-f8d}.
However, many approaches assume static goals, prior knowledge about collaborators, and full observability~\cite{wu2021too-589, ying2024goma-d5e}, limiting their applicability to settings where goals and responsibilities constantly change or the environment is only partially observable. 
Such settings are common in real-world collaboration, where agents must reason under uncertainty about both the task and their collaborators.

We address this challenge by developing an agent that maintains belief states over the environment and collaborators, enabling dynamic task allocation under partial observability, evolving tasks, and no prior knowledge of partners. 
Building on \citet{wu2021too-589}, our agent reasons about unobserved objects, updates task allocations from new evidence, and exchanges task-relevant information beyond direct observation with dialog. 
Together, these mechanisms support long-horizon collaborative planning under uncertainty about both the environment and other agents' behavior.

Our contributions are:
\begin{enumerate}
\item We introduce a language-lifted agent that maintains beliefs over unobserved state and incorporates natural-language communication as probabilistic evidence for task-allocation inference.
\item We introduce a novel framework for evaluating dynamic multi-agent task allocation with dynamically arriving tasks under both full and partial object-space observability, implemented within the Overcooked environment.
\end{enumerate}

\section{Related Works}\label{sec:related-works}

\subsection{Multi-Agent Planning}\label{sec:related-works:multi-agent}
Significant work in multi-agent planning has studied how agents coordinate task allocation with collaborators~\citep{stone2010ad,barrett2012learning,ribeiro2022assisting-d91}.
Ad hoc teamwork studies the setting where agents need to coordinate with previously unseen teammates without prior agreement on a joint policy~\citep{stone2010ad,barrett2012learning,ribeiro2022assisting-d91}.
\citet{wu2021too-589} introduce Bayesian Delegation to update task allocations in collaboration, which decomposes tasks into subtasks and uses inverse planning to infer a collaborator's latent subtask allocation. 
\citet{ying2024goma-d5e} extend this line to partially observable settings by combining recursive Theory of Mind reasoning with communication for human-agent collaboration.
However, these prior methods assume fixed goal specifications~\citep{stone2010ad, barrett2012learning, ribeiro2022assisting-d91, ying2024goma-d5e, wu2021too-589} and do not work when goals are continually evolving.
In contrast, our agent performs dynamic task allocation through natural-language dialog in partially observable environments without prior knowledge of the collaborator's behavior.

\subsection{Human-Agent Dialogue}\label{sec:related-works:human-robot-dialogue}
Researchers have long studied how to build collaborative agents that use dialogue~\citep{yu2025mixed-initiative-644,ying2024goma-d5e,liu2023llmpowered-786, Tellex2014AskingFH, gu2026continualrobotskilltask, ren2023robots}.
\citet{ijcai2018p1} describe dialogue as a form of interaction through which humans communicate task-relevant knowledge to embodied agents.
Prior works also develop agents that actively ask for help when the agent is stuck~\citep{Tellex2014AskingFH} or uncertain about the task~\citep{ren2023robots, gu2026continualrobotskilltask}.
Beyond help-seeking, recent work has incorporated dialogue into collaborative task reasoning, using language models to infer collaborators' mental states or generate coordinated actions~\citep{yu2025mixed-initiative-644,ying2024goma-d5e,liu2023llmpowered-786}.
In our setting, dialogue is used to communicate task-relevant information and provide evidence for deciding future actions.

\subsection{LLM-based Planning}\label{sec:related-works:language-guided-planning}
LLMs have been increasingly studied for generating plans and coordinating agents~\citep{zhang2023building-025, mandi2023roco-3ee, ahn2022do-9a9}.
\citet{liu2023llmpowered-786} propose an LLM-based hierarchical language agent for real-time human-AI coordination by combining language-based intention reasoning with reactive execution.
\citet{sun2025collabovercooked-f8d} show that LLM agents struggle with active collaboration and continuous adaptation by introducing a benchmark for evaluating LLM-based multi-agent systems on collaborative tasks.
Furthermore, \citet{valmeekam2023planning} show that LLMs perform poorly as autonomous long-horizon planners, but can provide useful heuristic guidance for other planners.
In our approach, the LLM is used to interpret dialogue as task-relevant evidence for an explicit multi-agent planning procedure.

% Our approach instead uses language to support a grounded multi-agent planning procedure: dialogue provides evidence about a collaborator's latent task allocation, while task selection remains grounded in observations, inferred collaborator behavior, and the currently available tasks. This separation matters in our setting, where new tasks arrive during execution and agents may observe different portions of the environment. We therefore investigate whether language can improve coordination when incorporated into an explicit task-allocation framework, rather than relying on an LLM alone to maintain and execute the collaborative plan.
\section{Methodology}\label{sec:method}

\begin{figure*}[t]
\centering
\begin{subfigure}[b]{0.48\linewidth}
    \centering
    \includegraphics[height=5cm]{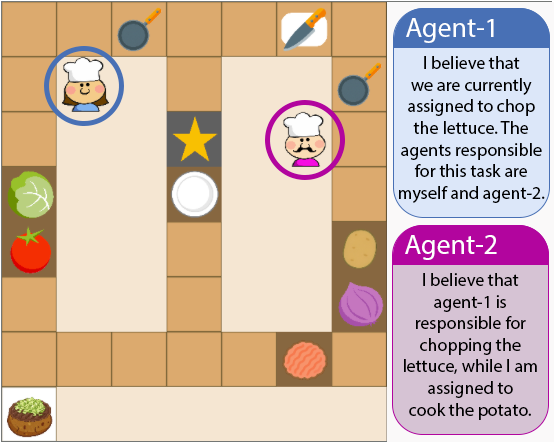}
    \subcaption{}
    \label{fig:method:language:comm-example}
\end{subfigure}
\hfill
\begin{subfigure}[b]{0.48\linewidth}
    \centering
    \includegraphics[height=5cm]{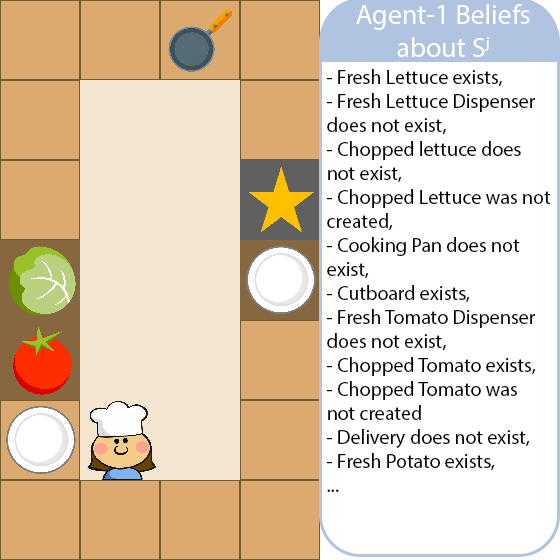}
    \subcaption{}
    \label{fig:method:partial-observe:belief}
\end{subfigure}
\caption{(a) An example of utterances between two agents. The agent circled in blue communicates its belief that both agents are currently collaborating to chop the lettuce. The agent circled in purple communicates its belief that it is cooking the potato while the other agent is chopping the lettuce. (b) Example of an agent's beliefs over the unobserved state space. In this example, the agent's latent propositions include object existence and object creation.}
\label{fig:method:combined}
\end{figure*}

\subsection{Preliminaries}\label{sec:method:prelim}
 
\citet{wu2021too-589} formulate collaborative task allocation as a multi-agent Markov decision process (MMDP), in which agents share a known static goal and infer which sub-task each collaborator is pursuing. 
Given an interaction history $H_{0:T}$ containing the states and joint actions observed from timestep $0$ through $T$, and letting $\mathbf{ta}$ denote the set of feasible task allocations, \citet{wu2021too-589} infer the most likely allocation as
\begin{equation}\label{eq:method:prelim:task-alloc}
ta^* = \arg\max_{ta \in \mathbf{ta}} P(ta \mid H_{0:T}).
\end{equation}
% This inference relies on a fully observed joint action, which does not hold under partial observability: an agent may not observe a collaborator's actions, its progress, or the portion of the environment in which it is operating, leaving behavior alone insufficient to reliably infer task allocation. This motivates our extension of task allocation to incorporate information communicated between agents, allowing them to reason about collaborators whose behavior cannot be directly observed.
In this formulation, the task allocation is computed with fully observed joint actions, which is not accessible when the collaborator's actions, progress, or environment are not visible.

\subsection{Decentralized POMDP}\label{sec:method:dec-pomdp}
 
To account for this partial observability, we extend the formulation of \citet{wu2021too-589} into a decentralized partially observable Markov decision process (\textit{Dec-POMDP}) with sub-tasks \cite{pynadath2002communicative-c4c}, defined by the tuple $<n, \mathcal{S}, \mathcal{A}_{1\dots n}, T, \mathbb{O}, \mathcal{O}, R, h, b_0, \mathcal{T}>$, where $n$, $\mathcal{S}$, $\mathcal{A}$, $T$, $R$, and $\mathcal{T}$ are defined by \citet{wu2021too-589}. The joint observation space is $\mathbb{O} = \mathbf{o}_1 \times \dots \times \mathbf{o}_n$, with $\mathcal{O}(\mathbf{o}\mid s)$ the probability of joint observation $\mathbf{o} \in \mathbb{O}$ given state $s$; the horizon $h$ is $t_{\max}$ timesteps completed; and $b_0 \in \Delta(\mathcal{S})$ is the initial state distribution at $t=0$.
 
\subsection{Language Lifted Planner}\label{sec:method:language}
  
To model dialogue interactions, we add communication to each agent's observation history. 
Each agent maintains a belief distribution over task allocations and broadcasts a message when this belief is sufficiently uncertain; Fig.~\ref{fig:method:language:comm-example} shows an example dialogue.
The agent estimates uncertainty over the current task allocation using Shannon entropy~\cite{shannon1948mathematical} and communicates when this entropy exceeds a predefined threshold; details are provided in Appendix~\ref{app:method:extended}.

% An entropy-based utterance policy controls the broadcasting: the agent communicates when the Shannon entropy of its task-allocation belief, computed by Equation~\ref{eq:method:language:entropy}, exceeds an $\epsilon$-fraction of the maximum possible entropy~\cite{shannon1948mathematical} (Appendix~\ref{app:method:extended}).

 % Each agent maintains its own belief distribution over task allocations and broadcasts a message when it is uncertain. 
 % As illustrated in Figure~\ref{fig:method:language:comm-example}, the broadcast action is decided by an entropy-based utterance policy: the agent communicates whenever the Shannon entropy of its task allocation belief exceeds an $\epsilon$-percentage threshold of the maximum possible entropy \cite{shannon1948mathematical} (Equation~\ref{eq:method:language:entropy}, Appendix~\ref{app:method:extended}). The utterance itself is generated by prompting an LLM with $ta^*$.

The observation history $H_{0:T}$ is a sequence of tuples containing the agent's observation $s_t$, joint action $\mathbf{a}_t$, and joint utterance $\mathbf{z}_t$ over all timesteps, denoted as $H_{0:T} = \{(\mathbf{z}_t, \mathbf{a}_t, s_t)\}_{t=0}^T$.
 % Agents include all broadcast utterances in their observation history, so  $H_{0:T} = \{(\mathbf{z}_t, \mathbf{a}_t, o_t)\}_{t=0}^T$, where $\mathbf{z}_t$ denotes the joint utterance at timestep $t$.
 % Equation~\ref{eq:method:prelim:task-alloc} is correspondingly extended to weight each allocation by both the joint action likelihood and a joint utterance likelihood term:
 We compute the posterior in Equation~\ref{eq:method:prelim:task-alloc} by weighting each task allocation according to both the observed action likelihood and utterance likelihood:
\begin{equation}\label{eq:method:language:language-posterior}
P(ta|H_{0:T}) \propto P(ta)\prod_{t=0}^{T}
P(\mathbf{a}_t \mid s_t, ta)\,
P(\mathbf{z}_t \mid ta).
\end{equation}
The terms $P(ta)$ and $P(\mathbf{a}_t \mid s_t, ta)$ are the task allocation prior and joint action likelihood, respectively, as defined by \citet{wu2021too-589}. 
The joint utterance likelihood $P(\mathbf{z}_t \mid ta)$ is computed directly from the LLM's output token probabilities for each agent's utterance. Details can be found in Appdx.~\ref{app:method:extended}.

\subsection{Partial Observability}\label{sec:method:partial-observe}
 
Our formulation can be extended into a partially observable setting.
Each agent maintains a belief state, and the interaction history becomes $H_{0:T} = \{(\mathbf{z}_t, \mathbf{a}_t, o_t, b_t)\}_{t=0}^T$, where unobserved actions are denoted $\emptyset$, $o_t$ is the agent's observation at timestep $t$, and $b_t$ is its belief over the unobserved state. 
To better approximate $Q$-values under partial observability, each agent maintains a belief over latent propositions $p$ describing traits of the unobserved portions of the environment (Figure~\ref{fig:method:partial-observe:belief}), updated at each timestep as:
\begin{equation}
\label{eq:method:partial-observe:belief}
P(p \mid o_T, H_{0:T-1}) \propto
\begin{cases}
\displaystyle \frac{\phi(p)}{b^j_{T-1}(p)}, & b^j_{T-1}(p) \neq 0,\\
0, &\text{otherwise},
\end{cases}
\end{equation}
where $\phi(p)$ combines evidence extracted from the current observation, the previous joint action $\mathbf{a}_{T-1}$, and the previous observation $o_{T-1}$ with the agent's inferred task allocation to estimate propositions that cannot be directly observed. The full derivation is provided in Appendix~\ref{app:method:extended}.
 
Accordingly, Equation~\ref{eq:method:language:language-posterior} is further extended with an observation likelihood term:

\begin{equation}
\label{eq:method:partial-observe:po-posterior}
\begin{aligned}
P(ta \mid H_{0:T}) &\propto P(ta)\prod_{t=0}^{T}
P(\mathbf{a}_t \mid o_t, b_t, ta)\,
P(\mathbf{z}_t \mid ta) \\
&\quad\times
P(o_t, \mid \mathbf{a}_{t-1}, o_{t-1}, b_{t-1}, ta),
\end{aligned}
\end{equation}

% For new results:
% \begin{equation}
% \label{eq:method:partial-observe:po-posterior}
% \begin{aligned}
% P(ta \mid H_{0:T}) &\propto P(ta)\prod_{t=0}^{T}
% P(\mathbf{a}_t \mid o_t, b_t, ta)\,
% P(\mathbf{z}_t \mid ta) \\
% &\quad\times
% P(o_t, b_{t}\mid \mathbf{a}_{t-1}, o_{t-1}, b_{t-1}, ta),
% \end{aligned}
% \end{equation}
% where $P(ta)$ is defined by \citet{wu2021too-589} and the utterance likelihood is defined by Eq.~\ref{eq:method:language:utterance-likelihood}. 
The details for the terms of this equation can be found in Appdx.~\ref{app:method:extended}.
% The joint action likelihood replaces unobserved actions with an exponentiated $Q$-value term (Equation~\ref{eq:method:partial-observe:joint-action}, Appdx.~\ref{app:method:extended}), and the observation likelihood approximates unobserved transitions using the inverse of the estimated value function (Equation~\ref{eq:observation-likelihood}, Appdx.~\ref{app:method:extended}).

\begin{figure*}[htbp]
\centering
\begin{subfigure}{0.48\linewidth}
\centering
\includegraphics[width=\linewidth]{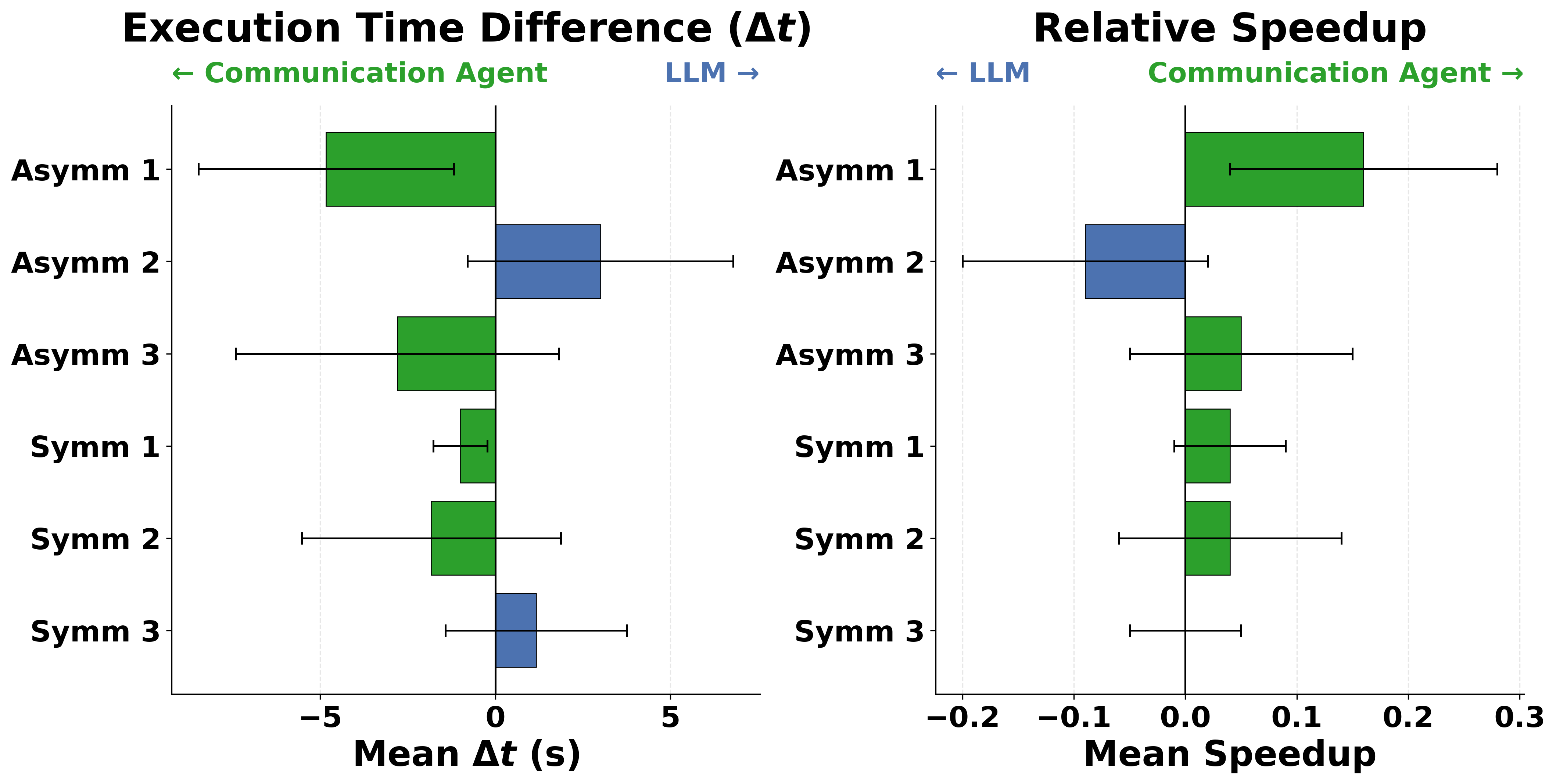}
\caption{Comm vs.\ LLM}
\label{fig:llm_comm_fo}
\end{subfigure}
\hfill
\begin{subfigure}{0.48\linewidth}
\centering
\includegraphics[width=\linewidth]{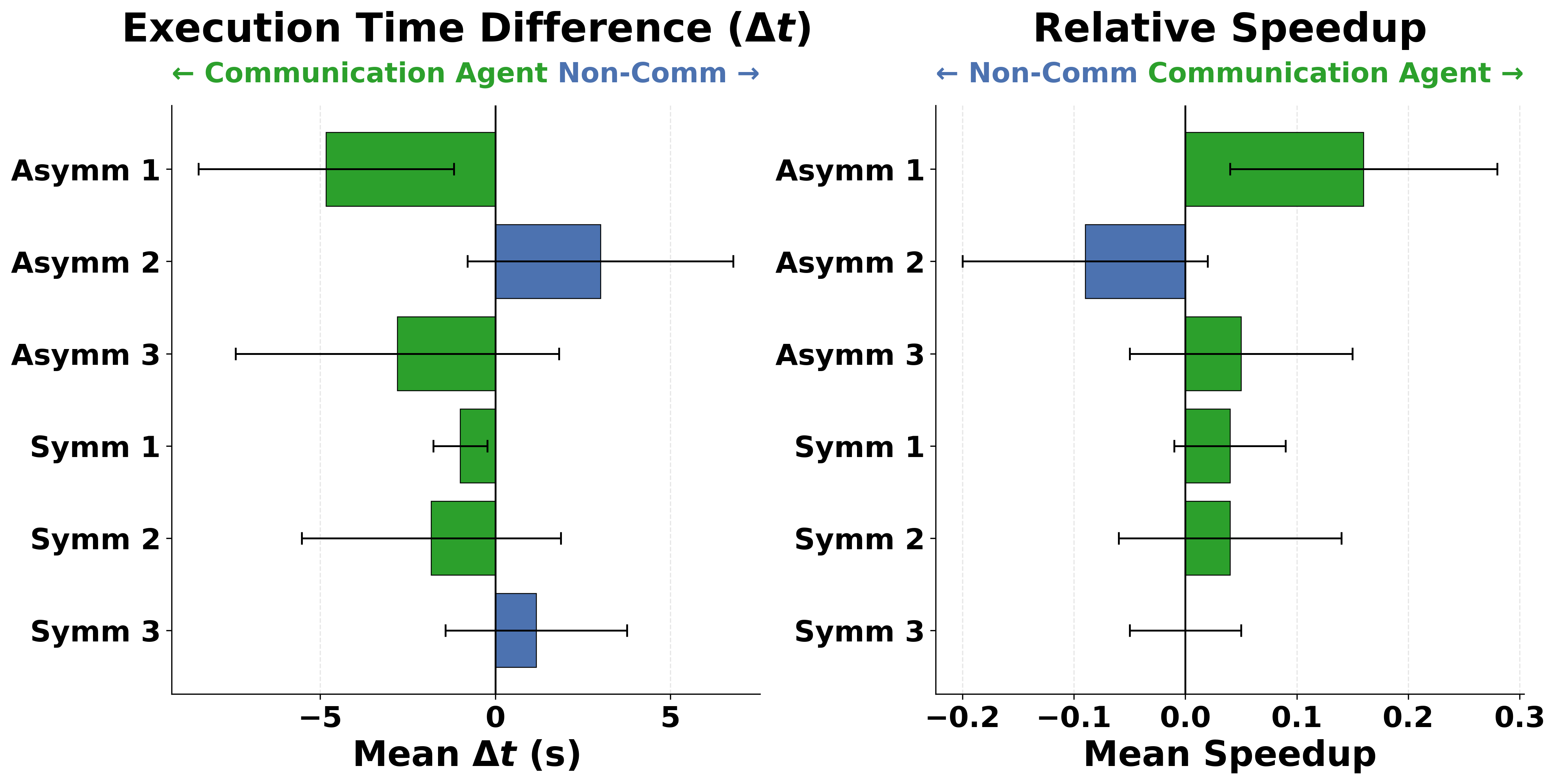}
\caption{Comm vs.\ Non-Comm}
\label{fig:noncomm_comm_fo}
\end{subfigure}
\caption{Communication agent performance relative to the two baselines under full observability. Error bars show standard error.}
\label{fig:results:fo}
\end{figure*}

\begin{figure*}[htbp]
\centering
\begin{subfigure}{0.48\linewidth}
\centering
\includegraphics[width=\linewidth]{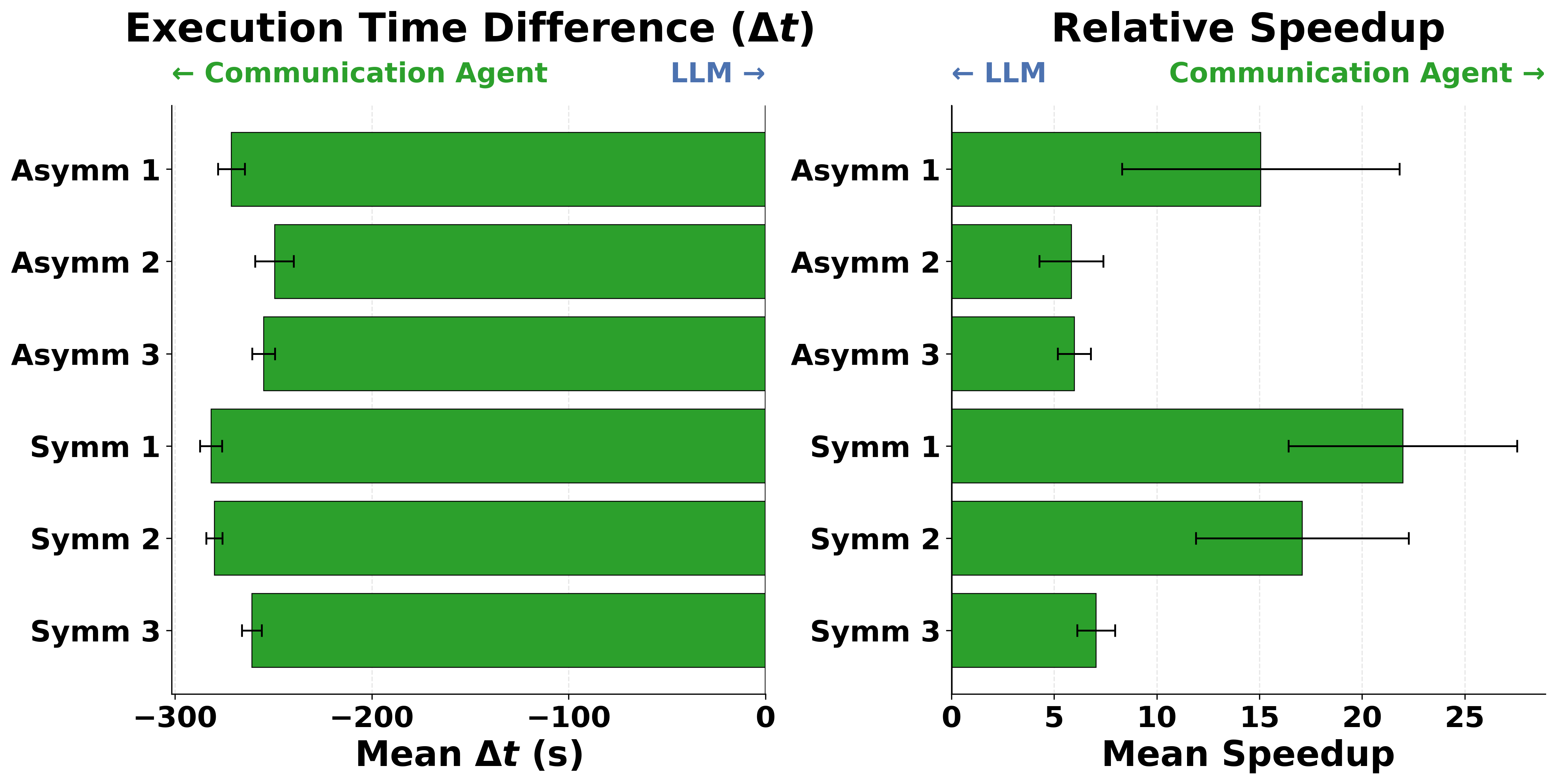}
\caption{Comm vs.\ LLM}
\label{fig:llm_comm_po}
\end{subfigure}
\hfill
\begin{subfigure}{0.48\linewidth}
\centering
\includegraphics[width=\linewidth]{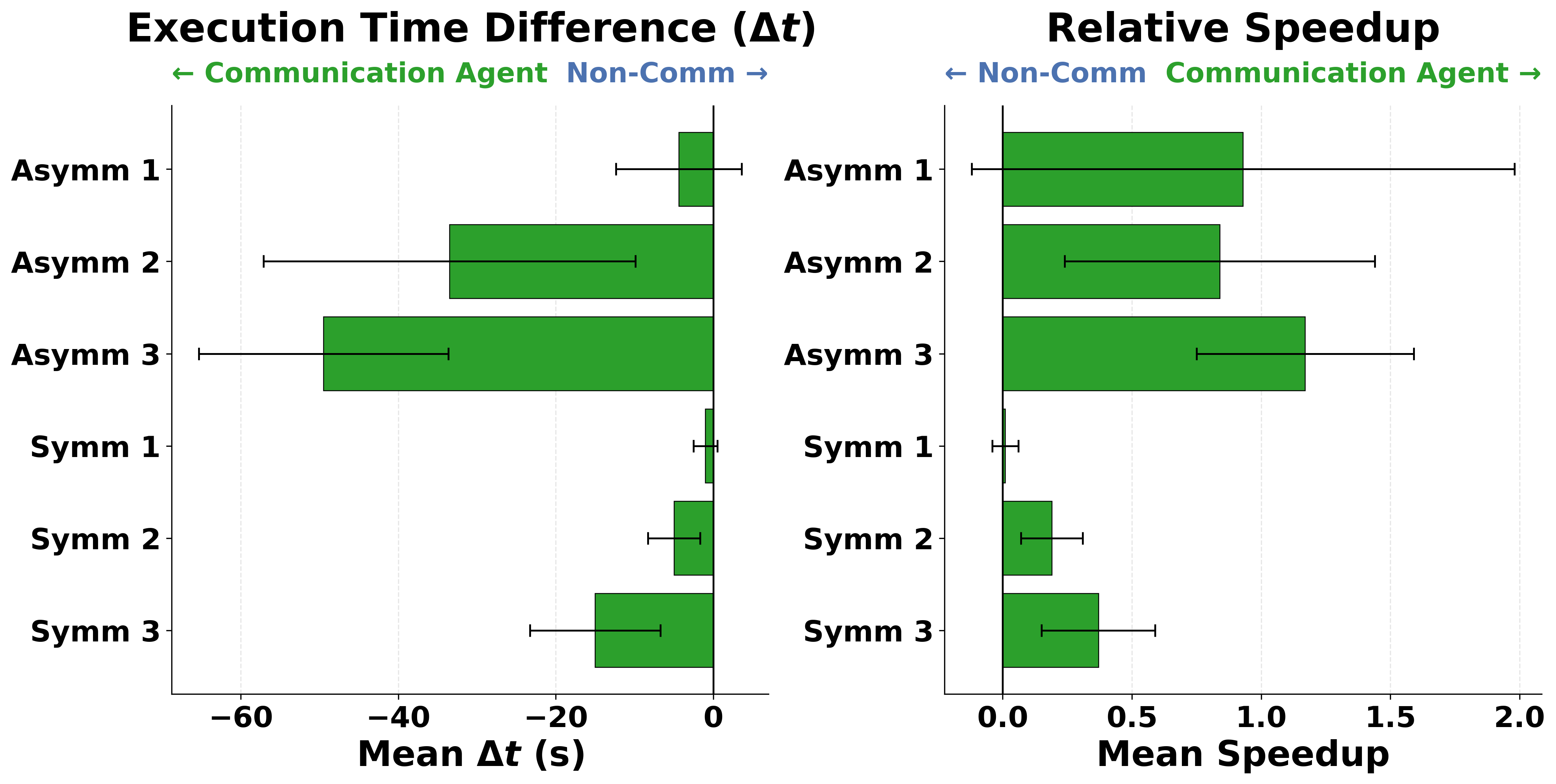}
\caption{Comm vs.\ Non-Comm}
\label{fig:noncomm_comm_po}
\end{subfigure}
\caption{Communication agent performance relative to the two baselines under partial observability. Error bars show standard error.}
\label{fig:results:po}
\end{figure*}
 
Together, these mechanisms allow the agent to perform dynamic task allocation under partial observability by maintaining explicit beliefs over both collaborator task allocation and the unobserved environment, while using dialogue to reduce uncertainty as tasks and goals evolve.
Detailed descriptions and derivations can be found in Appdx.~\ref{app:method:extended}.
% Together, these mechanisms produce an agent that maintains an explicit belief over its collaborator's task allocation and the unobserved environment, communicating proactively when that belief becomes too uncertain to act on (Equation~\ref{eq:method:language:entropy}, Appendix~\ref{app:method:extended}) and resolving uncertainty about the unobserved state through a belief update over latent propositions (Equation~\ref{eq:method:partial-observe:belief}). Unlike prior task-allocation frameworks, which infer allocation only from observable actions and a fully known state, or prior work on communication under partial observability, which typically assumes a static, predefined goal, our agent infers task allocation continually from the environment and interaction history, adapting its beliefs and communication as tasks and goals evolve.

\section{Results}\label{sec:results}
\subsection{Domain}\label{sec:results:domain}
We extend the \textit{Overcooked} environment introduced by \citet{wu2021too-589} with dynamic goal setting, where tasks are revealed incrementally during execution rather than fixed at the start.
We also introduce partially observable settings in which collaborators' actions or local environments are not accessible to the agent.
This environment allows us to test agents' reasoning and communication abilities on dynamic goals under different levels of uncertainty.
Additional details can be found in Appdx.~\ref{app:experiments}.

% We evaluate our approach in the \textit{Overcooked} domain, a cooperative multi-agent environment in which agents share a kitchen to prepare and deliver recipes, coordinating how work is distributed across ingredient dispensers, preparation stations, and counters. We extend the Overcooked environment introduced by \citet{wu2021too-589} with a dynamic goal-setting formulation: rather than a fixed task set known at the start of an episode, tasks are revealed incrementally throughout execution, requiring agents to continually reconsider task allocation as new goals appear (full task-arrival details in Appendix~\ref{app:experiments}). 

% We evaluate agents under both full and partial object-space observability (Figure~\ref{fig:observability}, Appendix~\ref{app:experiments}) to study dynamic task allocation under different levels of environmental knowledge.

\subsection{Baselines}\label{sec:results:baseline}
We compare against baselines suited to each observability setting. Under full observability, we adapt the task-allocation algorithm of \citet{wu2021too-589} to support dynamic task arrivals, modifying it to replan as the task queue changes over an episode. Under partial observability, a setting we introduce for this domain, we use an ablation of our own method with communication removed, isolating the contribution of explicit communication when agents lack full environmental knowledge. In both settings we also compare against a vanilla LLM agent, following the LLM-based agents benchmarked by \citet{sun2025collabovercooked-f8d}, which receives the same observability-appropriate information but selects actions directly, without the additional mechanisms in our methodology.
\subsection{Metrics}\label{sec:results:metrics}
We evaluate performance using timestep difference, $\Delta t = T_M - T_B$ (negative values favor the communication agent $M$ over baseline $B$), and speedup, $T_B/T_M - 1$ (positive values favor $M$), following \citet{ying2024goma-d5e}. We report the mean and standard error (SE) of each metric across runs.
\subsection{Experimental Results}\label{sec:results:experimental-results}
We evaluate full and partial observability separately, varying task queue length and layout within each setting; full experimental parameters are reported in Appendix~\ref{app:experiments}.

\subsubsection{Full Observability}
As shown in Fig.~\ref{fig:llm_comm_fo}, the communication agent consistently outperforms the vanilla LLM agent across both layouts, with the largest speedups at the queue length of one and diminishing gains as queue length increases. 
As shown in Fig.~\ref{fig:noncomm_comm_fo}, our agent is comparable to the non-communicative baseline across most conditions.
Because both agents observe the full object space, the non-communication agent already has access to much of the information communication would otherwise convey, limiting its added benefit.

\subsubsection{Partial Observability}
The communication agent consistently outperforms the vanilla LLM agent, with gains again largest at a queue length of one (Fig.~\ref{fig:llm_comm_po}).
However, communication now yields a clear advantage over the non-communication baseline (Fig.~\ref{fig:noncomm_comm_po}).
This advantage generally grows with queue length and is most pronounced in the asymmetric layout at a queue length of three. 
With less available knowledge, communication is more important to close the information gap between the agents.

\section{Discussion}\label{sec:discussion}

Our results show that explicit task allocation and communication provide structure that a vanilla LLM agent lacks: the LLM baseline fails to complete any configuration, while our agent succeeds across observability settings. 

The value of communication depends on observability. 
Under full observability, our agent performs similarly to the non-communication baseline because agents can infer collaborators' tasks from visible actions. 
Under partial observability, communication becomes increasingly important, especially as queue length grows, because actions alone no longer reveal enough about collaborators' tasks or progress.

\section{Limitations}\label{sec:limitations}
% Our communication mechanism is deliberately narrow. Utterances convey only task-allocation intent, not the granular beliefs, questions, or negotiation of human collaborators, likely omitting information especially useful under partial observability. Agents decide when to speak via an entropy-based trigger tied to their own uncertainty: a reasonable but likely suboptimal proxy, since human speakers also condition on what their partner already knows. More broadly, our constrained, task-oriented utterances do not capture the ambiguity, back-and-forth, or repair of human dialogue, and since all experiments pair our agent with itself, its interpretability to a novice human partner, a natural direction for future work, remains untested.
% Our experiments are also limited in scale: a single environment family, two layouts, task queue lengths of at most three, two baselines, and six runs per configuration, so results may not generalize to domains with different action spaces, longer horizons, larger teams, or more heterogeneous capabilities. Computational constraints also limited the extent of hyperparameter search, with values (Appendix~\ref{app:experiments}) fixed a priori rather than tuned per configuration. Finally, our formulation assumes a small, fixed number of agents, a known, fixed set of task types, and reliable, cost-free communication; real deployments may face delayed or lossy messages that contend with the resources needed for physical action, and violating these assumptions could degrade the coordination benefits reported here.
Our work has two main limitations. 
First, our communication mechanism is deliberately narrow: utterances convey task-allocation intent but not richer beliefs, questions, negotiation, or repair, and the entropy-based speaking trigger may not reflect what a collaborator already knows. 
Second, our experiments are limited in scale, with one environment family, two layouts, short task queues, two baselines, and six runs per configuration, and they assume known task types, a fixed agent set, and reliable cost-free communication.
\section{Conclusion}\label{sec:conclusion}
We introduce a Bayesian task allocation algorithm for dynamically arriving tasks and partial object-space observability, using language as probabilistic evidence within an explicit inference procedure. 
Our results show that this framework enables stronger dynamic task allocation under partial observability than the baselines, and that communication becomes more important as available information decreases.
% Future work should study larger teams, richer dialogue, human partners, and generalization beyond Overcooked.
% We introduced a language-lifted extension of Bayesian task allocation that lets agents coordinate under dynamically arriving tasks and partial object-space observability, treating communication as probabilistic evidence within an explicit inference procedure rather than delegating planning to an LLM's judgment. In an extended Overcooked domain, our agent consistently outperformed a vanilla LLM agent under both observability settings, matching a non-communication baseline under full observability and substantially outperforming it under partial observability -- indicating that inverse planning alone suffices when the environment is fully observable, while explicit communication becomes necessary as observability and task complexity grow. Future work includes larger teams, richer dialogue, human evaluation, and generalization beyond Overcooked.

\newpage
\bibliography{citations}
\appendix
\section*{Extended Metholodgy}\label{app:method:extended}
 
This appendix provides the full formulations for the utterance policy, the joint utterance likelihood, the belief update over latent propositions, and the partially observable likelihood terms introduced in Section~\ref{sec:method}.
 
\subsection{Utterance Policy}\label{app:method:extended:entropy}
 
An agent communicates whenever the Shannon entropy of its task allocation belief exceeds a fraction of the maximum possible entropy \cite{shannon1948mathematical}:
\begin{equation}
\label{eq:method:language:entropy}
-\sum_{ta \in \mathbf{ta}} P(ta \mid H_{0:T}) \ln P(ta \mid H_{0:T}) \geq \epsilon \ln |\mathbf{ta}|,
\end{equation}
where $\epsilon \in [0,1]$ controls the utterance threshold (Section~\ref{sec:method:language}).
 
\subsection{Utterance Likelihood}\label{app:method:extended:lang-posterior}
 
The joint utterance likelihood $P(\mathbf{z}_t \mid ta)$ used in Equation~\ref{eq:method:language:language-posterior} (Section~\ref{sec:method:language}) is computed as
\begin{equation}
\label{eq:method:language:utterance-likelihood}
\begin{aligned}
&P(\mathbf{z} \mid ta)\\
&\quad= \prod_{i:\mathcal{T}_i \in ta}
\begin{cases}
1, & \text{if } z_i = \emptyset,\\
\prod_{j=0}^{|z_i|-1} P(z_{i,j} \mid ta, z_{i,<j}), &\text{otherwise},
\end{cases}
\end{aligned}
\end{equation}
where $z_i = \emptyset$ indicates that agent $i$ did not produce an utterance. In the latter case, the token probabilities are obtained directly from the output logits of the LLM.
 
\subsection{Belief Update Derivation}\label{app:belief}
 
This subsection provides the full derivation of the belief update introduced in Section~\ref{sec:method:partial-observe} (Equation~\ref{eq:method:partial-observe:belief}), extending the formulation from Sections~\ref{sec:method:prelim} and~\ref{sec:method:language} to environments with partially observable object spaces. An agent cannot directly observe the complete state of the environment and must reason about objects that may exist outside its field of view; each agent therefore maintains probabilistic beliefs over latent propositions describing the hidden region of the environment, recursively updated using both environmental observations and communication from its collaborator.
 
\subsubsection{Belief Representation and Initialization}\label{app:belief:representation}
 
We partition the object space into three regions: $S^i$, the region observable to agent $i$; $S^j$, the region observable only to agent $j$; and $S^s$, the shared divider region visible to both agents. Since the contents of $S^j$ are not observable, agent $i$ maintains beliefs over propositions describing its state. A proposition $p$ is a Boolean statement describing a latent property of $S^j$ — for example, whether an ingredient or action object exists within $S^j$, or whether an object was created within $S^j$.
 
At each timestep $T$, agent $i$ maintains a belief distribution over every proposition $p$, defined as the posterior probability
\begin{equation}
b_T^j(p)=
\begin{cases}
P(p \mid o_T) & \text{if } T = 0,\\
P(p \mid o_T, H_{0:T-1}) & \text{otherwise},
\end{cases}
\label{eq:belief-definition}
\end{equation}
where $o_T$ denotes the observation at timestep $T$, and $H_{0:T-1}=\left\{(\mathbf{z}_t,a_t^i,o_t,b_t^j)\right\}_{t=0}^{T-1}$ is the interaction history consisting of communicated utterances $\mathbf{z}_t$, agent $i$'s actions $a_t^i$, observations $o_t$, and beliefs $b_t^j$ from previous timesteps.
 
The initial beliefs are constructed from two environment assumptions: every task is completable, and every ingredient initially present is associated with a dispenser. Consequently, any object that is not directly observed must either lie within $S^j$ or have an unknown location. For each ingredient $x_m \in X$, the initial belief over its presence in the hidden region is
\begin{equation*}
\begin{aligned}
b_0^j(x_m)&=P(x_m \in S_0^j \mid o_0)\\
&=\begin{cases}
1&\text{if }x_m \in X_0\setminus(S_0^i\cup S_0^s),\\
0&\text{if }x_m \notin X_0,\\
b_{\text{init}} &\text{otherwise}.
\end{cases}
\end{aligned}
\end{equation*}
Since no actions have yet been performed at the initial timestep, the probability that any object was created within the hidden region is initialized to zero,
\begin{equation*}
b_0^j\left(C(x_m,S_T^j)>C(x_m,S_{T-1}^j)\right)=0.
\end{equation*}
Similarly, for every ingredient initially present, the belief over the location of its corresponding dispenser is
\begin{equation*}
\begin{aligned}
b_0^j(d(x_0))&=P(d(x_0)\in S_0^j \mid o_0)\\
&=\begin{cases}
1&\text{if }d(x_0)\notin(S_0^i\cup S_0^s),\\
0.5 &\text{otherwise}.
\end{cases}
\end{aligned}
\end{equation*}
and beliefs over action objects are initialized the same way: for every action object $y \in Y$,
\begin{equation*}
b_0^j(y)=P(y\in S_0^j \mid o_0)=\begin{cases}
1&\text{if }y\notin(S_0^i\cup S_0^s),\\
0.5 \textbf{}&\text{otherwise}.
\end{cases}
\end{equation*}
 
\subsubsection{Posterior Belief Update}\label{app:belief:posterior}
 
At every subsequent timestep, agent $i$ updates its beliefs using newly observed evidence and the utterance produced by agent $j$. Since agent $i$ has complete knowledge of its own actions and their deterministic effects, any change in the observation that its own action cannot explain must have resulted from agent $j$, and therefore provides evidence about $S^j$. We isolate this information with an evidence extraction function $E(o_T,a^i_{T-1},o_{T-1}) = e$, where $e$ is obtained by comparing the expected observation after executing $a^i_{T-1}$ with the actual observation $o_T$.
 
The belief over every proposition is updated by combining two complementary sources of information: the likelihood of the proposition given the observed evidence, and the likelihood of the proposition implied by the communicated task allocation. The posterior belief is
\begin{equation}
\begin{aligned}
&P(p|o_T,H_{0:T-1}) \propto\\
&\begin{cases}
\dfrac{\begin{array}{@{}c@{}}
P(p|e,o_{T-1},H_{0:T-2})\\[2pt]
{}\times\sum\limits_{ta\in\mathbf{ta}}P(z^j_{T-1}|ta)\\[2pt]
{}\times P(p|ta,o_{T-1},H_{0:T-2})
\end{array}}
{b^j_{T-1}(p)}
&
\text{if }b^j_{T-1}(p)\neq0,
\\[1.5em]
0
&
\text{otherwise},
\end{cases}
\end{aligned}
\label{eq:belief-posterior}
\end{equation}
where $P(p|e,o_{T-1},H_{0:T-2})$ is the proposition likelihood conditioned on the extracted evidence, $P(z^j_{T-1}|ta)$ is the likelihood of the communicated utterance under a candidate task allocation, $P(p|ta,o_{T-1},H_{0:T-2})$ is the proposition likelihood implied by the task allocation, and $b^j_{T-1}(p)$ is the prior belief from the previous timestep.
 
\subsubsection{Evidence-Conditioned Likelihood}\label{app:belief:evidence}
 
The observation likelihood is grounded in the deterministic dynamics of the environment: since only agent $j$ can explain changes inconsistent with agent $i$'s own action, each observed state transition provides evidence about the hidden environment. We categorize this evidence into four types — an object was picked up, set down, delivered, or no observable evidence occurred — and each type updates only the propositions directly affected by the transition, leaving all remaining propositions at their previous belief.
 
When an object disappears from the observable region without being removed by agent $i$, agent $i$ infers that agent $j$ picked up the object within $S^j$; the picked-up object must therefore have existed within the hidden region immediately before the action, and no object could have been created during the same timestep:
\begin{equation}
\begin{aligned}
&P(p|e_{\mathrm{pick\ up}},o_{T-1},H_{0:T-2})\\
&\quad=
\begin{cases}
1 & \begin{array}[t]{@{}l@{}}\text{if }p\text{ states that the picked-up}\\\text{object exists,}\end{array}\\[2pt]
0 & \text{if }p\text{ is a creation proposition,}\\
b^j_{T-1}(p) & \text{otherwise}.
\end{cases}
\end{aligned}
\label{eq:prop-evidence-pick-up-likelihood}
\end{equation}
 
A newly observed object indicates that agent $j$ produced the object within the hidden region before placing it into the observable space. Because the environment dynamics are known, the agent can infer the ingredients, action objects, and dispensers required to construct the newly observed object. Let $X_x$ denote the intermediate objects consumed during construction, $Y_x$ the action objects required by the recipe, $D_x$ the dispensers for ingredients not directly observed being collected, and $X'_x$ the objects that could subsequently be produced from the newly observed object:
\begin{equation}
\begin{aligned}
&P(p|e_{\mathrm{set\ down}},o_{T-1},H_{0:T-2})\\
&\quad=
\begin{cases}
1 & \text{if }p\text{ is about }y,\;y\in Y_x,\\
1 & \text{if }p\text{ is about }d(x),\;d(x)\in D_x,\\
0 & \begin{array}[t]{@{}l@{}}\text{if }p\text{ states that the set-down}\\\text{object exists,}\end{array}\\[2pt]
0 & \text{if }p\text{ is about }x,\;x\in X_x\cup X'_x,\\
0 & \text{if }p\text{ is a creation proposition},\\
b^j_{T-1}(p) & \text{otherwise}.
\end{cases}
\end{aligned}
\label{eq:prop-evidence-set-down-likelihood}
\end{equation}
 
Delivery evidence is treated analogously to set-down evidence: delivering an object reveals the recipe that must have been executed within the hidden region to produce it, so the same inferred ingredients, action objects, and dispensers update the proposition likelihood:
\begin{equation}
\begin{aligned}
&P(p|e_{\mathrm{deliver}},o_{T-1},H_{0:T-2})\\
&\quad=
\begin{cases}
1 & \text{if }p\text{ is about }y,\;y\in Y_x,\\
1 & \text{if }p\text{ is about }d(x),\;d(x)\in D_x,\\
0 & \begin{array}[t]{@{}l@{}}\text{if }p\text{ states that the delivered}\\\text{object exists,}\end{array}\\[2pt]
0 & \text{if }p\text{ is about }x,\;x\in X_x\cup X'_x,\\
0 & \text{if }p\text{ is a creation proposition},\\
b^j_{T-1}(p) & \text{otherwise},
\end{cases}
\end{aligned}
\label{eq:prop-evidence-deliver-likelihood}
\end{equation}
and when no evidence is observed, every proposition retains its previous belief:
\begin{equation}
P(p|e_{\mathrm{none}},o_{T-1},H_{0:T-2})
=
b^j_{T-1}(p).
\label{eq:prop-evidence-none-likelihood}
\end{equation}
 
\subsubsection{Task-Allocation-Conditioned Likelihood}\label{app:belief:task-alloc}
 
While environmental evidence provides information about actions that have already occurred, communication provides information about actions an agent intends to perform in the future. Because completing a task requires the necessary prerequisite objects, an utterance describing a task allocation implicitly conveys information about the hidden environment: for example, if agent $j$ communicates that it will independently prepare a dish, agent $i$ can infer that the ingredients, action objects, and dispensers required for that task are available within $S^j$.
 
As in Section~\ref{sec:method:language}, we estimate the likelihood of an utterance under a candidate task allocation using the output logits of a large language model,
\begin{equation}
P(z^j_{T-1}\mid ta)
=
\prod_{l=0}^{|z^j_{T-1}|}
P(z_l\mid ta,z_{<l}),
\label{eq:utterance-likelihood-prop}
\end{equation}
where $z_l$ denotes the $l$-th generated token and $z_{<l}$ the preceding tokens.
 
Given a task allocation, agent $i$ can infer the objects that must already exist within the hidden region for the communicated plan to be feasible; because the environment dynamics are known, these prerequisite objects also reveal the intermediate objects, action objects, and dispensers required to produce them. Let $X_{ta}$ denote the intermediate objects consumed while constructing the prerequisite objects implied by the task allocation, $Y_{ta}$ the corresponding action objects, and $D_{ta}$ the dispensers associated with ingredients not directly observed being collected:
\begin{equation}
\begin{aligned}
&P(p|ta,o_{T-1},H_{0:T-2})\\
&\quad=
\begin{cases}
1 & \text{if }p\text{ is about }y,\;y\in Y_{ta},\\
1 & \text{if }p\text{ is about }d(x),\;d(x)\in D_{ta},\\
1 & \begin{array}[t]{@{}l@{}}\text{if }p\text{ is about }x\text{ and }x\text{ is a}\\\text{prerequisite for }ta,\end{array}\\[2pt]
0 & \text{if }p\text{ is about }x,\;x\in X_{ta},\\
0 & \text{if }p\text{ is a creation proposition},\\
b^j_{T-1}(p) & \text{otherwise}.
\end{cases}
\end{aligned}
\label{eq:prop-ta-likelihood}
\end{equation}
This likelihood captures the information conveyed by communication independently of the environmental observations; during the posterior update, it is combined with the observation likelihood to produce a belief distribution that reflects both the agent's observed behavior and its communicated intentions.
 
\subsection{Partially Observable Likelihood Terms}\label{app:method:extended:po-terms}
 
Under partial observability (Section~\ref{sec:method:partial-observe}), the joint action likelihood in Equation~\ref{eq:method:partial-observe:po-posterior} is defined as
\begin{equation}
\label{eq:method:partial-observe:joint-action}
\begin{aligned}
&P(\mathbf{a}_t|o_t, b_t, ta)\\
&\quad= \prod_{i: \mathcal{T}_i \in ta}
\begin{cases}
1 &\text{if } a^i_t = \emptyset,\\
\exp(\beta Q^*_{\mathcal{T}_i}(o_t, b_t, a^i_t)) &\text{otherwise},
\end{cases}
\end{aligned}
\end{equation}
where $a^i_t = \emptyset$ signifies that agent $i$'s action at timestep $t$ is unobserved; the latter case is defined by \citet{wu2021too-589}.
 
The observation likelihood is defined as
\begin{multline}
\label{eq:observation-likelihood}
P(o_t \mid b_{t}, \mathbf{a}_{t-1}, o_{t-1}, ta) \\
= \prod_{i:\mathcal{T}_i \in ta}
\begin{cases}
1, & \text{if } a_t^i \neq \emptyset,\\
\dfrac{1}{V^*_{\mathcal{T}_i}(o_t,b_{t-1})}, & \text{otherwise.}
\end{cases}
\end{multline}
If an agent's action is directly observed ($a_t^i \neq \emptyset$), that component of the observation is treated as fully explained by the task allocation and assigned likelihood $1$. Otherwise, the likelihood is approximated using the inverse of the estimated value function under task $\mathcal{T}_i$, reflecting the intuition that observations corresponding to higher-value states are more likely under the agent's latent task.

% New formulation for next set of results:
% \begin{multline}
% \label{eq:observation-likelihood}
% P(o_t, b_{t} \mid , \mathbf{a}_{t-1}, o_{t-1}, b_{t-1}, ta) \\
% \propto \prod_{i:\mathcal{T}_i \in ta}
% \begin{cases}
% 1, & \text{if } a_t^i \neq \emptyset,\\
% \exp(\beta C^*_{\mathcal{T}_i}(o_t,b_{t})), & \text{otherwise.}
% \end{cases}
% \end{multline}
\section*{Extended Results}\label{app:experiments}
\begin{figure}[H]
\centering
\begin{subfigure}{0.48\linewidth}
\centering
\includegraphics[width=\linewidth]{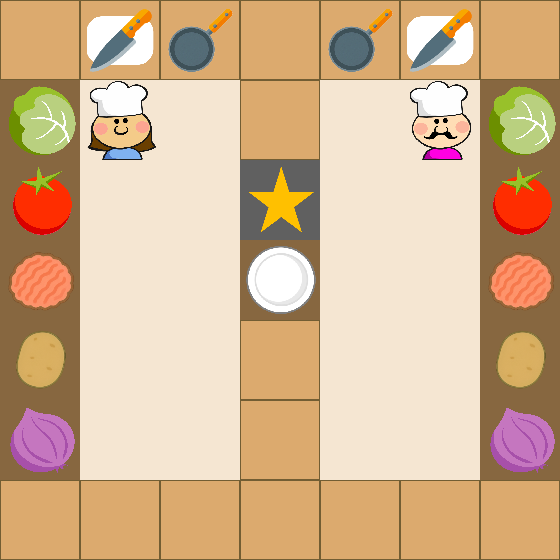}
\caption{Symmetric Layout}
\label{fig:results:experiment:symm}
\end{subfigure}
\hfill
\begin{subfigure}{0.48\linewidth}
\centering
\includegraphics[width=\linewidth]{images/asymm.png}
\caption{Asymmetric Layout}
\label{fig:results:experiment:asymm}
\end{subfigure}
\caption{(a) Each agent has access to all action objects and fresh ingredient dispensers, allowing experiments with agents of even capabilities. (b) Each agent has access to different action objects and a fresh ingredient dispenser, allowing experiments with uneven capabilities.}
\label{fig:results:experiment:experiment-layouts}
\end{figure}

\paragraph{Task arrival.} Agents are tasked with completing $N$ tasks within a horizon of $t_{\max}$ timesteps. Tasks are revealed incrementally, and agents do not know the complete task set in advance, although they know the set of possible task types. At each timestep, a new task is revealed with probability $r$ until all $N$ tasks have been revealed. To maintain continuous availability of work, at least one task remains active in the task queue until all tasks have been completed.
\begin{figure}[htbp]
\centering
\begin{subfigure}[b]{0.48\linewidth}
\centering
\includegraphics[width=\linewidth]{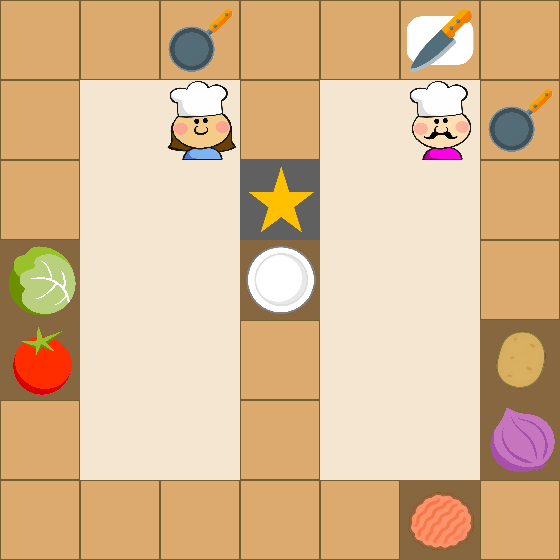}
\caption{Fully Observable}
\label{fig:results:domain:fully-observable}
\end{subfigure}
\hfill
\begin{subfigure}[b]{0.48\linewidth}
\centering
\includegraphics[width=\linewidth]{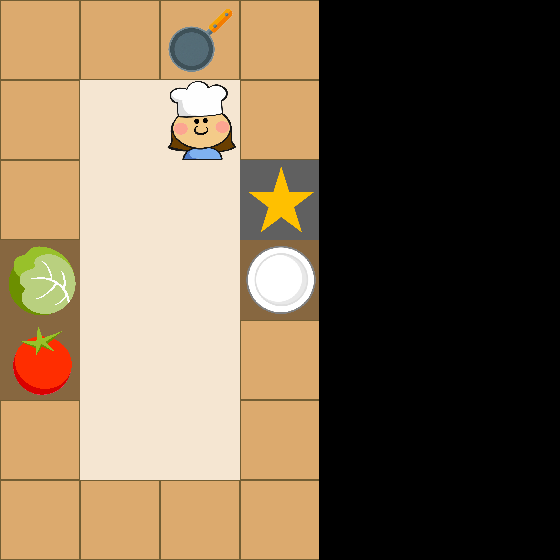}
\caption{Partially Observable}
\label{fig:results:domain:partially-observable}
\end{subfigure}
\caption[Different Observabilities]{Agent~1's observation of the same environment state under full and partial observability. The task queue is displayed at the bottom. Under full observability, Agent~1 observes the entire object space, including Agent~2. Under partial observability, Agent~1 observes only its partition and the shared counter space.}
\label{fig:observability}
\end{figure}
\paragraph{Experimental setup.} We evaluate task queue lengths of $1$, $2$, and $3$ on both the symmetric (Figure~\ref{fig:results:experiment:symm}) and asymmetric (Figure~\ref{fig:results:experiment:asymm}) layouts, under both full and partial observability. Tasks are randomly sampled from six task types. Each experimental variation is run six times, with each run using a different sampled task sequence, and each run is capped at $300$ timesteps.
\paragraph{Hyperparameters.} All experiments use the hyperparameters $\beta = 1.3$, $\epsilon = 0.7$, $r = 0.025$, $\alpha = 0.01$, and $\tau = 2$. We adopt $\beta = 1.3$ from \citet{wu2021too-589}. We set $r = 0.025$ to reduce the likelihood that new tasks appear at the beginning of each timestep, while the remaining hyperparameters are fixed across all experiments.

% Accidentally used parameters for new set of exeperiment: $\beta = 1.3$, $\epsilon = 0.5$, $r = 0.025$, $\alpha = 0.01$, and $\tau = 2$

\end{document}